# Domain Adaptation from Synthesis to Reality in Single-model Detector for Video Smoke Detection

Gao Xu[a], Yongming Zhang[a], Qixing Zhang[a,*], Gaohua Lin[a], Jinjun Wang[a]

[a] State Key Laboratory of Fire Science, University of Science and Technology of China, Hefei 230026, China

*Corresponding author: Qixing Zhang, Email: qixing@ustc.edu.cn, Tel: 86-551-63600770

**Abstract**

This paper proposes a method for video smoke detection using synthetic smoke samples. The virtual data can automatically offer precise and rich annotated samples. However, the learning of smoke representations will be hurt by the appearance gap between real and synthetic smoke samples. The existed researches mainly work on the adaptation to samples extracted from original annotated samples. These methods take the object detection and domain adaptation as two independent parts. To train a strong detector with rich synthetic samples, we construct the adaptation to the detection layer of state-of-the-art single-model detectors (SSD and MS-CNN). The training procedure is an end-to-end stage. The classification, location and adaptation are combined in the learning. The performance of the proposed model surpasses the original baseline in our experiments. Meanwhile, our results show that the detectors based on the adversarial adaptation are superior to the detectors based on the discrepancy adaptation. Code will be made publicly available on http://smoke.ustc.edu.cn.

# 1. Introduction

Object detection has received great attention during recent years. Smoke detection, as a type of object detection, is a challenging sub-problem that has attracted massive attention in the intelligent video surveillance for wildfire prevention. Yuan [1] proposed a double mapping framework to extract multi-scale partitions features for video smoke detection. A higher order linear dynamical system (h-LDS) descriptor [2] is applied to smoke detection based on the higher order decomposition of the multidimensional smoke image data. [3] used wavelets and support vector machines to characterize smoke. Three-dimensional spatiotemporal features are extracted from smoke frames to generate bag-of-features (BoF) [4] for smoke verification. However, the problems existed in video smoke detection have trailed its application, including lack of abundant samples and efficient detection algorithms, while great process has been made in general object detection, e.g. pedestrian detection. The advances in object detection produce powerful baseline systems, such as the one-stage detector MSCNN [5], YOLO [6], SSD [7], and two-stage detector Faster R-CNN [8], FPN [9], R-FCN [10]. Inspired by recent progress in computer graphics, it is more available to use synthetic images [11, 12] to probe the ability of detectors. Due to the gap between the synthetic and real smoke image data, training on synthetic smoke data hinders the power of the state-of-the-art detectors.

As the different distributions between real and synthetic smoke samples, the model trained on synthetic smoke samples performs poorly when applied to real scenes. Domain adaptation aims to shift the model trained from source domain to the target domain. Recently, there has been some works on domain adaptation for pedestrian detection [13-16]. In brain decoding studies, a sparse-coded cross-domain adaptation approach [17] was proposed to transfer the knowledge learned from visual domain to brain domain. The work in [18] investigated the possibility of automatically ranking source CNNs prior to utilize them in the given target task. In order to better learn domain-invariant features, [19] developed a two-dimensional subspace alignment approach based on 2D principal component analysis to better adapt convolution activations.

The work [14] firstly demonstrated adaptation of virtual and real worlds for developing an object detector, it used the pyramidal sliding window to obtain positives and negatives, and trained a domain adapted pedestrian classifier on the samples from the two domains. We provided the first attempt to apply the domain adaptation to smoke classification in images [20]. However, all the relevant works above take the object detection and domain adaptation as two independent parts. In this paper, we construct an end-to-end detector based on the state-of-the-art single-model detectors (SSD and MS-CNN) to combine domain adaptation and detection.

## 2. Background

In general, the deep networks for detection are used to position the objects and then classify the candidate regions for detail categories, such as pose estimate, face recognition. Meanwhile, as the lack of suitable annotated training images (e.g. wildfire smoke images), researchers [11] have trained the detector on the synthetic data with ground truth annotations. As the appearance gap between synthetic and real data will degrade the performance of system, [21, 22] applied domain adaptation with the classifier of the detector. However they focused on optimizing the representation of object region, namely setting the domain mixer with classifier on the bottom of feature extractor on object region. [23] showed the domain shift factors including spatial location accuracy, appearance diversity, image quality and aspect distribution affect the performance of the detectors. As it is impossible to eliminate these factors between synthetic and real data, domain adaptation is indispensable to the end-to-end detection architecture.

For video smoke detection, as the complexity of environment and lacking of initial wildfire smoke samples, we use the synthetic smoke images to train the detector. Our system needs to combine the detection and domain adaptation. The domain adaptation confuse the features extracted from synthetic and real smoke, preventing the poor performance due to their different distribution. [24] proposed a gradient reversal layer for a domain mixer, which acts an identity transform during the forward propagation and multiplies the gradient from the subsequent level by a negative during the back propagation. [25] set a bottleneck layer for adaptation to learn a representation that minimizes the distance between the source and target distributions, preventing overfitting to the particular nuances of the source distribution. In [26], they optimized the architecture for domain invariance to facilitate domain transfer and use a soft label distribution matching loss to transfer information between tasks.

[27] played the first attempt to combine detection and pose estimation at the same level based on SSD, which is faster and more accurate than the object detection pipeline adding the pose estimation as a secondary classification stage based on the resulting detection. Inspired by the ideas from the method proposed by [26], which choose an adversarial loss to minimize domain shift through simultaneously training domain classifier and domain mixer iteratively, we add the domain prediction branches to the state-of-the-art single-model baselines (SSD and MS-CNN), training the location, classification and domain confusion iteratively. The structural design and training procedures are different between these two baselines. In addition, we try the discrepancy adaptation methods for minimizing the difference between domains to rebuild the overall objective loss in the baseline.

## 3. Overview of the single-model detector

In this section, we introduce the detection baseline constructed with domain adaptation. Our method is a generic solution for detector to learn generalizable representations of objects from the synthetic data and it performs well when applied to real scenes. To achieve this, the single-model detectors SSD and MS-CNN are chosen as baseline. Different from the concrete samples for the general classification network based on domain adaptation, the one-stage detector process a large set of candidate object proposals that densely cover spatial positions, scales, and aspects ratios regularly sampled across an image [28].

### 3.1. Single shot multibox detector

SSD is built by adding convolutional feature layers to the end of the truncated base network. It shares locations for multiple categories and these added layers predict locations and categories using a set of convolutional filters. Based on the multi-scale feature maps of the convolutional layers following the base network, the convolutional predictors output a set of category and location prediction at the corresponding scale. All the candidate locations are assembled and sampled by the matching strategy in the overall loss layer. SSD samples the negatives by hard negative mining to control the ratio between the negatives and positives at most 3:1.

### 3.2. Multi-scale CNN

MS-CNN is similar to the SSD but consist of two sub-networks. The network detects objects through several detection branches in the proposal sub-network. Each detection branch predicts location and categories simultaneously through convolutional layers. Different from SSD, the detection branches are independent from each other in sampling work. Meanwhile, the classification loss is re-weighted in the detection layer to prevent the damage when no positive training samples are available in this branch. To cover objects well, the object detection sub-network is added to the MS-CNN. The proposals sampled from the proposal network are used to extract features of object in a deconvolution layer.

## 4. Domain adaptation for the detector

To demonstrate the simplicity and effectiveness of our methods, we make the modifications to the original systems of [5, 7] with corresponding adaptation to their architecture. In the following, we construct the adaptation inside SSD and MS-CNN for smoke detection.

### 4.1. Adversarial Adaptation

The current deep networks for object detection, whether relying on region proposal or not, will choose the positive and negative samples based on the location and confidence prediction during training procedure. Assumed that the positives (objects) are recognized relatively well, then domain adaptation are applied to adapt the feature representation of these samples to make them similar between the source domain and target domain. As the predicted annotated samples may contain the false positives which may influence the data distribution, [16] design an element-wise multiply layer for the decomposed convolutional layer trained by the abundant source samples, then set an unsupervised regularization followed to adapt the true target samples to some extent. The element-wise multiply layer acts similarly to the filters in detector for extracting class and location information. We hypothesized that this operation takes the classification confidence as the class information of predicted region and adapt it by the unsupervised regularization. In the end-to-end network without region proposal, the last feature extraction layer is not generally fully connected layer but convolution layer. We transform the element-wise multiply layer to operate on the last convolution layer, but it is difficult to converge in the results.

Particularly, in the case we considered, there are abundant annotated synthetic smoke samples of scene-specific and limited annotated real smoke samples. In the following, we add domain outputs through the convolution prediction layers at each scale in SSD and MS-CNN. It should be noted that predictions share locations for multiple categories in SSD, while MS-CNN predicts location individually between categories.

[27] observed that sharing outputs for pose across all the object categories performs better than having separate pose outputs for each object category. We will share domain outputs by adding a series of branches. Inspired by the adversarial training framework [26], we jointly train the classification, location and domain adaptation. We have tried to adapt the representation for the whole images, but it performs badly. Namely, the distributions of the whole images and object-region can't be adapted in the same term. It confirms our point of view that the proposal regions are taken to domain adaptation in pedestrian detection in [13]. To learn a detection representation that is domain invariant for the object-region, we propose to rely the adaptation on the classification and location prediction.

#### 4.1.1. Adversarial adaptation for SSD

In the general classification network based on domain adaptation, the classifier and domain mixer are independent structures after the feature extraction layer. In our method, we add a series of domain prediction by adding $3 \times 3$ filters to transform each feature map to predict domain scores. In our case, each location in feature layer will output classification scores, localization offsets, and domain prediction scores like [27]. However, the accuracy of the domain prediction is based on the trustworthiness of the classification and location predictions. We firstly train the model $M^{pre}$ of baseline SSD with annotated samples $(X^{real}+X^{render}, Y^{render}+Y^{real})$ and obtain the believable proposals of positives (smoke) and negatives (background), as well as the retrained feature representation $\theta_{repre}$, $\theta_{loc\_conf}$. Then we use the pre-trained model $M^{pre}$ to initialize the model $M$ of modified detector with domain prediction,

$$L_{domain\_classifier}(\theta_{repre}, \theta_{loc\_conf}; \theta_{domain}) = -\sum_{i} \log p_k$$

Where $p_k$ is the softmax of the domain classifier.

With the fixed parameters of feature extraction layers, we train a strong domain classifier $M_{dc}$ for the detector. To learn a fully domain invariant representation, we fix the weights $\theta_{domain}$ of the domain branch and upgrade the $\theta_{repre}$ with the poor domain classifier,

$$L_{domain\_mixer}(\theta_{loc\_conf}, \theta_{domain}; \theta_{repre}) = -\sum_{i}[0.5 * \log p_k + 0.5 * \log(1 - p_k)]$$

Then we obtain the domain mixer. For a particular feature representation and domain representation, to retrain the detector for optimizing the objectives as following,

$$L_{multibox}(\theta_{domain}, \theta_{repre}; \theta_{loc\_conf}) = \frac{1}{N}(L_{conf}(x,c) + \alpha L_{loc}(x,l,g))$$

### 4.1.2. Domain Adaptation for MS-CNN

On the other hand, the original training of MS-CNN is adopted with a two-stage procedure. The first stage is to train the trunk CNN layers. The resulting model is used to initialize the second stage, the detection sub-network (up-sampled by deconvolution to detect small objects) is trained with the matched proposal predicted from the trunk CNN layers. In the original sub-network of the MS-CNN, each branch emanating from different layers of the trunk share the location prediction with classes. We add convolution layers follow each detection layer to predict domain label. The learning is initialized with the VGG-Net. We modifies the original two-stage training procedure to three-stage training procedure. The first stage trains the model $M$ on the two domain with annotated samples $(X^{real}, Y^{real})$ and $(X^{render}, Y^{render})$ with a small trade-off coefficient $\lambda$, meanwhile, the parameters in each branches of the trunk layers are optimized with the parameters of conv1_1 to conv4_3 layers fixed. The second stage optimize the parameters in the whole trunk layers except for the parameters of the domain branch fixed. Then the two stages perform iterative updates for a domain-invariant representation $M$ in the trunk layer. The resulting model is used to initialize the third stage. The third stage optimize the sub-network $M^d$ of the MS-CNN with the fixed $M$.

For each detection layer,

$$l(X,Y|W) = L_{cls}(p(X),y) + \lambda[y \geq 1]L_{loc}(b,\hat{b}) + \eta[y \geq 1]L_{domain}$$

Where $p(X)$ means weighting the positives and negatives to restrain the bias caused by that no positives training samples are available for a detection layer. The computation of $L_{domain}$ is the iterative update of the domain classifier and domain mixer, similar to the method in SSD.

The total loss is as follows,

$$\varsigma(W, W_d) = \sum_{m=1}^{M}\sum_{i \in S^m} \alpha_m l^m(X_i, Y_i | W) + \sum_{i \in S^{M+1}} \alpha_{M+1} l^{M+1}(X_i, Y_i | W, W_d) + \sum_{i \in S^{M+1}} \alpha_{M+1} l_{domain}(X_{i,real}, X_{i,render})$$

Where $l_{domain}(X_{i,real}, X_{i,render})$ achieves the adaptation in the detection layer of the sub-network.

### 4.2. Discrepancy Adaptation

Except for the adversarial loss for domain adaptation mentioned above, the discrepancy adaptation methods for minimizing the difference between two domains like MMD (maximum mean discrepancy) [29], CORAL (correlation alignment) [30] and Euclidean distance loss are widely used as well. Obviously, compared to the methods based on adversarial loss [24, 26], this kind of domain adaptation methods don't need the additional branches for domain prediction. It is important to note that the numbers of proposals sampled from the two domains are different from each other. MMD (maximum mean discrepancy) is the most widely used unsupervised method [16, 25] at present, whose original formula is as follows,

$$L_{MMD} = \left\| \frac{1}{|X_S|} \sum_{x \varepsilon X_s} \Phi(X_s) - \frac{1}{X_T} \sum_{x \varepsilon X_T} \Phi(X_t) \right\|_F^2$$

There are no guarantees about the alignment of classes between each domain in the calculation of similar loss. Because the negatives with large number determine the loss. Inspired by [31], we reweight the positives and negatives by their number,

$$L_{MMD} = \left\| \left[ \frac{1}{N_p} \sum_{x \in p} \Phi(X_S) + \frac{1}{N_n} \sum_{x \in n} \Phi(X_S) \right] - \left[ \frac{1}{N_p} \sum_{x \in p} \Phi(X_T) + \frac{1}{N_n} \sum_{x \in n} \Phi(X_T) \right] \right\|_F^2$$

In order to investigate the effect of the general similarity measurements to the adaptation in detector, we implement the adaptation based on Euclidean loss and correlation alignment in SSD and MS-CNN. The correlation alignment [30] aims to align the second-order statistics of the source and target distributions with a linear transformation.

$$L_{euclidean} = \frac{1}{2N} \sum_{i=1}^{N} \left\| x_i^{source} - x_i^{target} \right\|_2^2$$

$$L_{coral} = \frac{1}{4d^2} \left\| C_{source} - C_{target} \right\|_F^2$$

Where N is set as the common number of matched smoke proposals. $C_S$ and $C_T$ are covariance matrices of source and target feature representations of $d$-dimension.

As these measurements need two inputs in same dimensions, namely, it needs the same number of proposals in two domain. We take the litter number of proposal boxes of the two domains as the common number. (e.g. the numbers of matched smoke proposals in synthetic and real domains are 300 and 400 respectively, the top 300 of matched smoke proposals in real domain are chosen to compute the loss).

## 5. Experiments on smoke detection

Due to the dispersion of smoke, it ranges from thick to light while spreading out. It is difficult to give a unique bounding box for smoke in the test set. Compared with the general objects, smoke does not have a fixed shape, as both the local and the whole can be used as object of smoke. Therefore, the feature maps at different scales can give predicted boxes with high confidence score. When the detector runs on the test set, it may give several boxes for an image with a ray of smoke. So we propose a measurement that based on the segmented region of smoke. We synthesized the smoke images using Mitsuba [32]. The synthetic smoke and real smoke images are different on a pixel level to some extent but similar in term of the shape and contour.

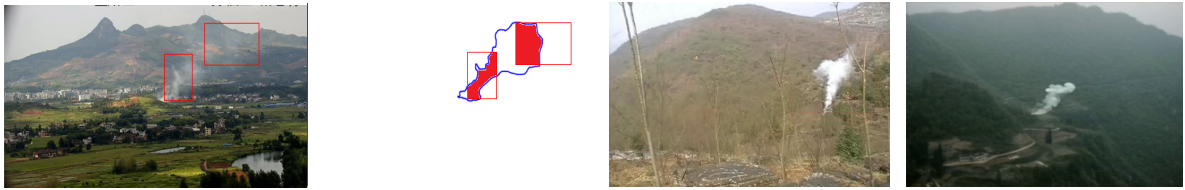

Figure 1: From left to right: detection box in test image, overlap between detection box and segmentation of smoke, real training image and synthetic training image.

Detections are considered true or false positives based on the area of overlap with ground truth bounding boxes [33]. The general area of overlap $\alpha_0$ between the predicted bounding box $B_p$ and ground truth bounding box $B_{gt}$ is formulated by:

$$\alpha_0 = \frac{area(B_p \cap B_{gt})}{area(B_p \cup B_{gt})}$$

By contrast, we define the overlap (predicted boxes with ground truth region, red in Figure 1.) as:

$$\alpha_{our} = \frac{\sum_i^n B_{map} \cap B_i}{\sum_i^n B_i}$$

Where, A is the ground truth region of smoke, B is the predicted box with confidence above confidence threshold. We use this overlap $\alpha_{our}$ to measure the location of the detector.

**5.1. Evaluation**

The test result shows that there are mainly two types of errors in smoke detection: confusion detection (low confidence below the threshold or detection on backgrounds) and missing detection (the number n of B with confidence above threshold is 0).

It is worth noting that [6] showed that different detectors perform differently in localization errors and background errors. It proposed that YOLO performances well when applied to person detection in artwork as it learns generalizable representations of objects. They explained that these non-region-based detectors models the size and shape of objects, as well as relationships between objects and where objects commonly appear.

Table 1: Detection results ( peak $\alpha_{our}$ ) of different cases

|  | Original baseline | | Constructed by adaptation |
| --- | --- | --- | --- |
|  | Real smoke images | Mixed dataset | Source (synthetic) + Target (real) |
| SSD | 31.90 | 47.40 | 56.90 |
| MS-CNN | 38.26 | 48.54 | 51.91 |

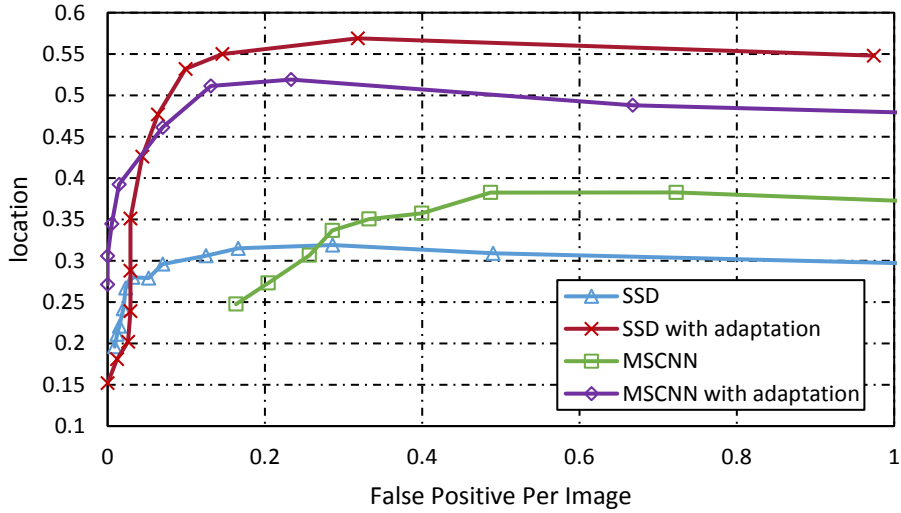

Figure 2: The average location accuracy of SSD and MS-CNN in the test set with varying confidence threshold.

As shown in Fig. 2, the performance of the baseline constructed by adaptation surpass the original baseline, while the confidence threshold gradually increasing. Table.1 represents the detection results of different cases.

## 5.2. Ablation Study

In this section, we draw upon several measures to analysis the mechanism of adaptation in the detector.

### 5.2.1. Should negative samples be abandoned?

In general domain adaptation classification networks, there is a balanced[7] sample distribution in all categories in both source dataset and target dataset. As a contrast, the state-of-the-art detectors control the sampling of positives and negatives. In general, the positives are chosen through the overlap between anchor bounding box and ground truth box, with the rest (background and low overlap) selected by designated sampling strategy. The statistic distributions of our samples are different from the general case in domain adaptation.

Moreover, the distribution of objects $S_+^m$ and non-objects $S_-^m$ of the training samples $S^m$ is heavily asymmetric. Namely, the proportion $\gamma$ is high in $|S_-| = \gamma |S_+|$. SSD and MS-CNN adopt the sampling strategy to compensate for this imbalance to control the $\gamma$. This bias in prior distributions of the training set will result in degraded domain adaptation performance [31]. In our experiment, the methods based on discrepancy adaptation perform not well except for the correlation alignment. We will analysis the reason in the following. Fig. 3 represents the results of the sampling heuristics alternatives about whether to abandon negatives in adaptation constructed to the baseline SSD.

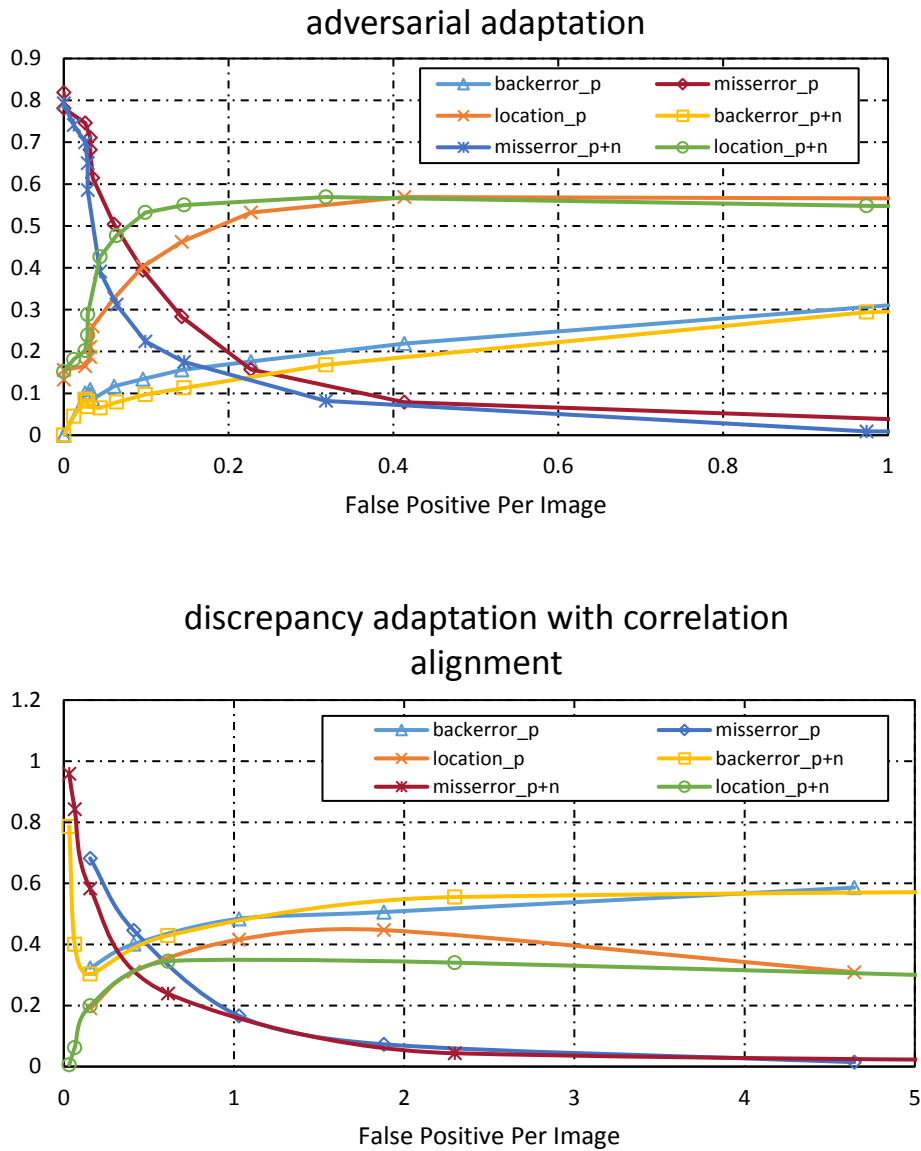

Figure 3: The background error, miss detection and location accuracy of detector with different sampling heuristics for adaptation.

The baseline adapted using adversarial loss with the fixed foreground-to-background ratio in its sampling heuristics has higher location accuracy and lower error than the alternative with the negatives abandoned. In contrast, the baseline adapted using correlation alignment with the negatives abandoned has higher location accuracy and lower missing detection than the alternative to sample backgrounds. Moreover, the confidence threshold for best performance of the latter is higher than the former. According to our analysis, the negative sampling is important for the methods based on correlation alignment.

**5.2.2. How is the distributions of proposals sampled?**

Experiments show that the performance of the SSD and MS-CNN based on adversarial adaptation surpasses the original baseline. In contrast, these detectors based on the discrepancy adaptation perform poorly. As the goal of our method is to adapt the distributions of smoke-proposals from the two domains during training procedure. Fig 4 shows the statistical distributions of proposals sampled from the two domains in the training procedure of the methods based on Euclidean distance and correlation alignment. The numbers of smoke object proposals sampled from real and synthetic images of the method based on correlation alignment are more balanced overall than the method based on Euclidean distance. What is more remarkable is that the mean value of the real proposals sampled is easy to worth far more than the mean value of the synthetic proposals sampled.

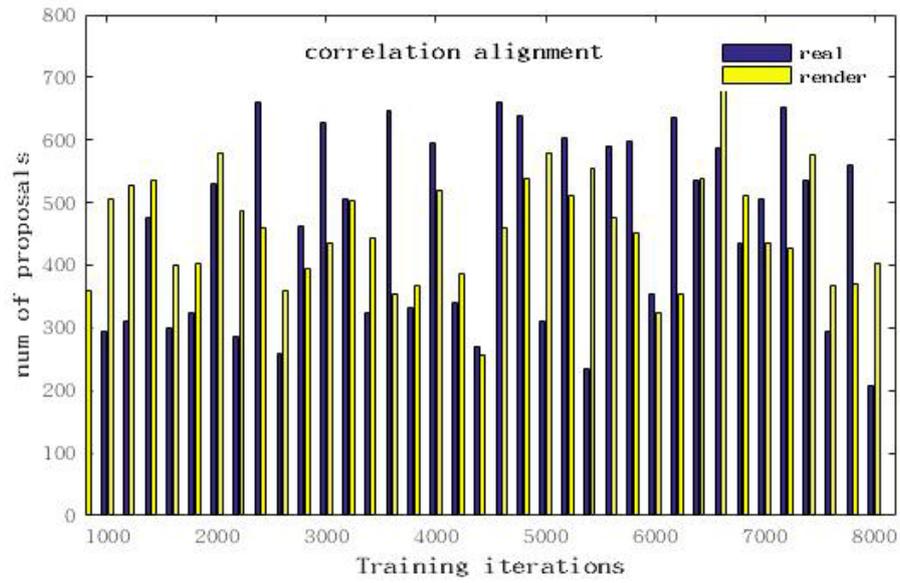
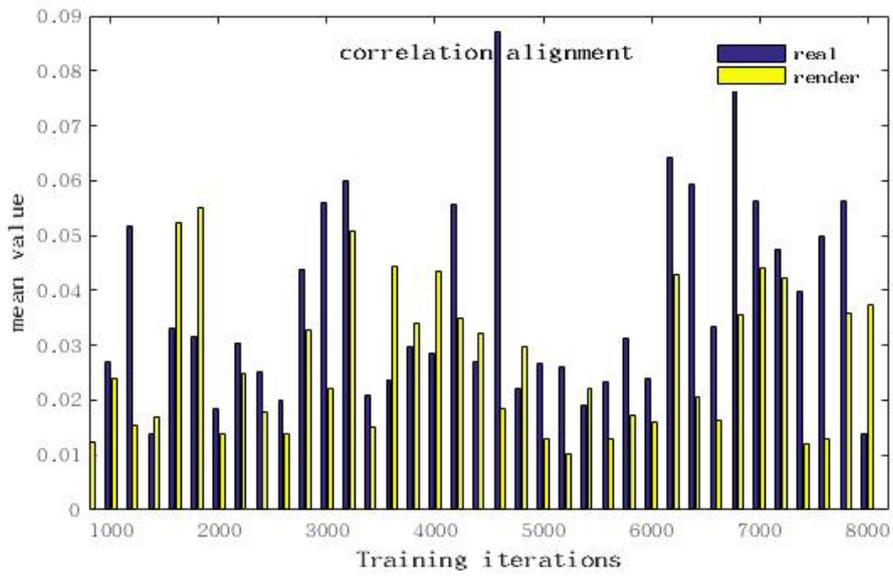
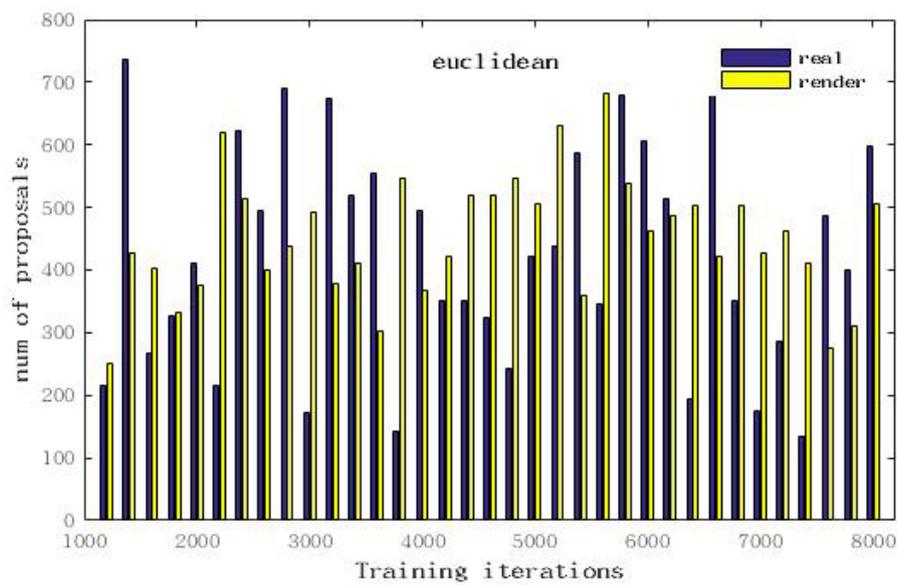

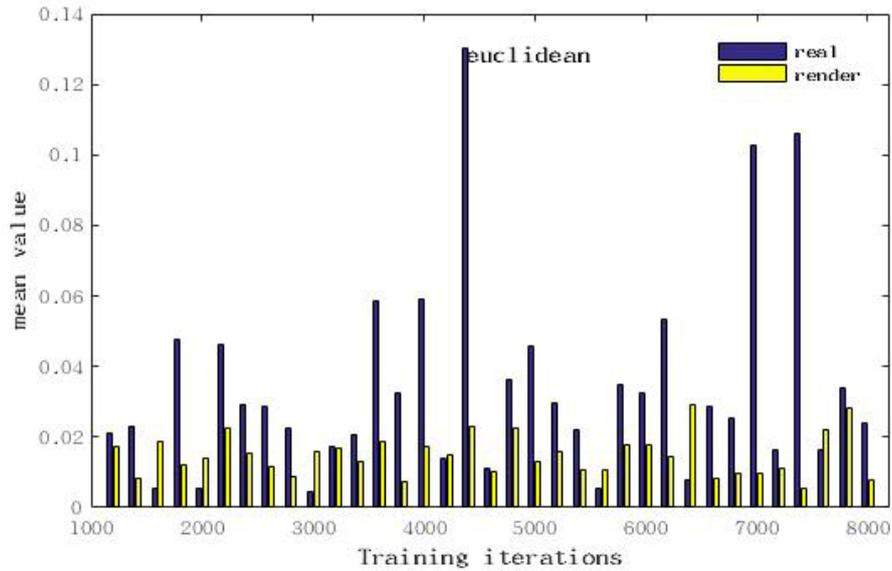

Figure 4: Comparison of the numbers and mean values of the smoke object proposals sampled from the two domains during training procedure.

**5.2.3. Error analysis**

Fig 5 represents the confusion detection and missing detection of the baseline based on the adversarial loss. In the confidence threshold 0.2 for the best location accuracy, the error of baseline adapted is approximate to the original SSD. If we pursue lower background false positives rate, the corresponding false negative rate will rise. Unfortunately, it is easier for the baseline adapted to cause more missing detection.

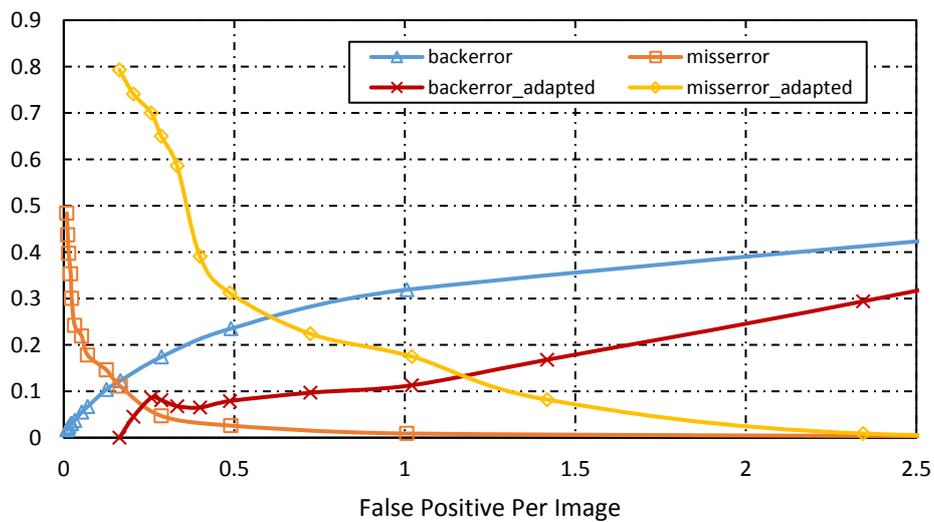

Figure 5: Errors analysis of different methods in confusion and missing.

**5.2.4. Multiple adaptation at different scales**

As deep features eventually transition from general to specific along the network, the feature transferability drops significantly in higher layers with increasing domain discrepancy. In classification network, generally only the last task-specific feature extraction layer is adapted. [34] proposed a network with all the layers corresponding to task-specially feature (fc6-fc8) adapted. In their words, multi-layer adaptation can make the task-specific layers transferable and jointly adapting the representation layer and the classifier layer. In our detector, feature maps with different scales are adapted through domain branches in the overall objective loss. Since the domain branch added to the lower layers affects their gradients more than the other detection branches. We test the architectures SSD with domain branches designed to validate its contribution. Fig 6 shows that the baseline with adaptation branches set at base-network layer fc7 and conv6_2 demonstrates the promising results. As the model detect the small smoke region in image using the information from shallow layers, it is essential to transfer the information from the feature map in low level. Meanwhile, the adaptation for feature map with different scales may cause overfitting to the particular nuances of the source distribution, more fine work will be done in this aspect.

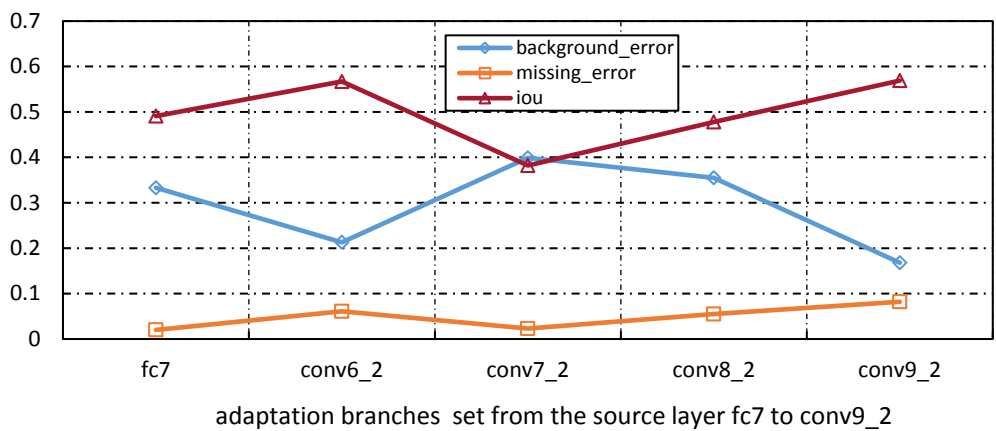

Figure 6: Different detection results using multiple adaptation.

## 6. Conclusions

We have proposed a framework based on the state-of-the-art single-model detectors for smoke detection. With the abundant annotated synthetic smoke samples and limited real smoke samples, our model combine the domain adaptation and object detection in each branch of detection layer. The performance of our methods show significant improvement over the original baseline. This paper provides a new method for the application of synthetic samples for object detection. The baseline SSD and MS-CNN are end-to-end single-model detectors, we plan to evaluate our methods in the region-based detectors like Faster-RCNN, F-RCN.

## Acknowledgements

This work was supported by the National Key Research and Development Plan (Grant No. 2016YFC0800100), the Anhui Provincial Key Research and Development Program (Grant No. 1704a0902030), and the Fundamental Research Funds for the Central Universities (Grant No. WK2320000033 and No. WK2320000032). The authors gratefully acknowledge all of these supports.